%% file: arxiv.tex
\documentclass[10pt,twocolumn,letterpaper]{article}

\usepackage{mysty}
\usepackage{cvpr}
\usepackage{times}
\usepackage{epsfig}
\usepackage{graphicx}
\usepackage{amsmath}
\usepackage{amssymb}

\usepackage{multirow}
\usepackage{booktabs}
\usepackage{siunitx}
\usepackage{subcaption}
\usepackage{bigstrut}
\usepackage{float}

\usepackage[breaklinks=true,bookmarks=false]{hyperref}

\cvprfinalcopy 

\def\cvprPaperID{3329} 

\begin{document}

\title{Comprehension-guided referring expressions}

\author{Ruotian Luo \\
TTI-Chicago\\
{\tt\small rluo@ttic.edu}
\and
Gregory Shakhnarovich\\
TTI-Chicago\\
{\tt\small greg@ttic.edu}
}

\maketitle

\input{abstract_intro}

\input{related}

\input{models}

\input{methods}

\input{experiments}

\input{conclusion}

\input{acknowledge}

{\small
\bibliographystyle{unsrt}
\bibliographystyle{ieee}
\bibliography{ref,ref1}
}

\end{document}

%% file: abstract_intro.tex
\begin{abstract}
We consider generation and comprehension of natural language referring
expression for objects in an image. Unlike generic ``image captioning"
which lacks natural standard evaluation criteria, quality of a
referring expression may be measured by the receiver's ability to
correctly infer which object is being described. Following this
intuition, we propose two approaches to utilize models trained for
comprehension task to generate better expressions. First, we use a
comprehension module trained on human-generated expressions, as a
``critic" of referring expression generator. The
comprehension module serves as a differentiable proxy of human
evaluation, providing training signal to the generation
module. Second, we use the comprehension module in a
generate-and-rerank pipeline, which chooses from candidate expressions
generated by a model according to their performance on the
comprehension task. We show that both approaches lead to improved
referring expression generation on multiple benchmark
datasets.

\end{abstract}

\section{Introduction}
Image captioning, defined broadly as automatic generation of text
describing images, has seen much recent attention. Deep
learning, and in particular recurrent neural networks (RNNs), have led
to a significant improvement in state of the art. However, the
metrics currently used to evaluate image captioning are mostly
borrowed from machine translation. This misses the naturally multi-modal distribution of
appropriate captions for many scenes.

Referring expressions are a special case of image captions. Such
expressions describe an object or region in the image, with the goal
of identifying it uniquely to a listener. Thus, in contrast to generic
captioning, referring expression generation has a well defined
evaluation metric: it should describe an object/region so that human can easily comprehend the description and find the location of the object being described.

In this paper, we consider two related tasks. One is the comprehension
task (called natural language object retrieval in \cite{Hu2015}),
namely localizing an object in an image given a referring
expression. The other is the generation task: generating a
discriminative referring expression for an object in an image. Most
prior works address both tasks by building a sequence generation model. 
Such a model can
be used discriminatively for the comprehension task, by inferring the region which maximizes the expression posterior.

\begin{figure}[t]
\centering
\par\vspace{10pt}
\includegraphics[width=0.45\textwidth]{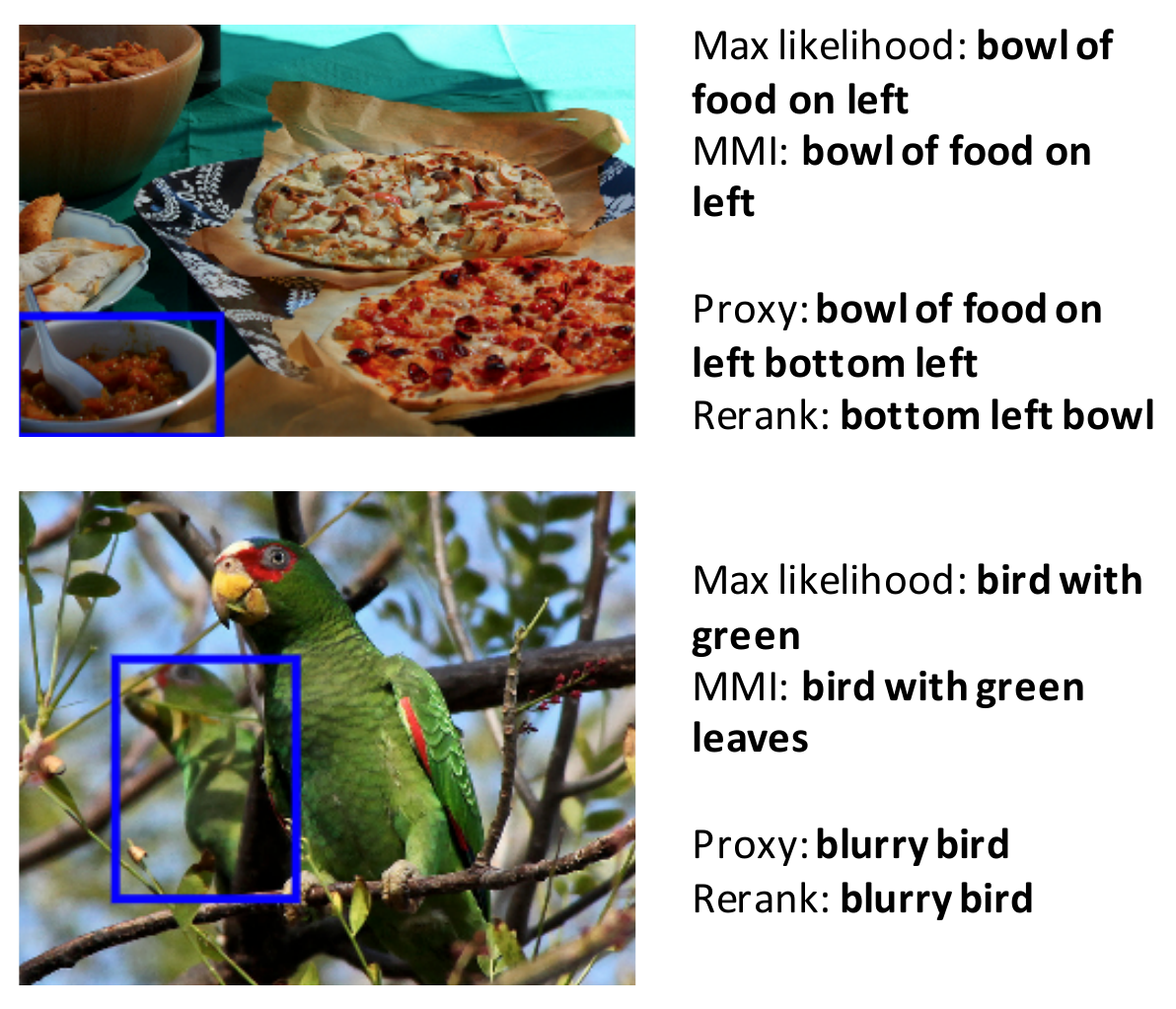}
\caption{These are two examples for referring expression generation. For each image, the top two expressions are generated by baseline models proposed in \cite{Mao2016a}; the bottom two expressions are generated by our methods.}
\label{fig:first}
\end{figure}

We depart from this paradigm, and draw inspiration from the generator-discriminator structure
in Generative Adversarial
Networks\cite{Goodfellow2014,Radford2015}. In GANs, the generator
module tries
to generate a signal (e.g., natural image), and the discriminator module
tries to tell real images apart from the generated ones. For our task,
the generator produces referring expressions. We would like these
expressions to be both intelligible/fluent and unambiguous to
human. Fluency can be encouraged by using the standard
cross entropy loss with respect to human-generated
expressions). On the other hand, we adopt a comprehension model as the ``discriminator" which tells if the expression can be correctly dereferenced. Note that we can also regard the comprehension model as a ``critic" of the ``action" made by the generator where the ``action" is each generated word. 

Instead of an adversarial relationship between the two modules in GANs,
our architecture is explicitly collaborative -- the comprehension module
``tells'' the generator how to improve the expressions it produces. Our methods are much simpler than GANs as it avoids the alternating
optimization strategy -- the comprehension model is separately trained
on ground truth data and then fixed. To achieve this, we adapt the
comprehension model so it becomes differentiable with respect to the
expression input. Thus we turn it into a proxy for human understanding
that can provide training signal for the generator.

Thus, our main contribution is the first (to our knowledge)
attempt to integrate automatic referring expression generation with a
discriminative comprehension model in a collaborative framework.

Specifically there are two ways that we utilize the comprehension
model. The {\bf generate-and-rerank} method uses comprehension on the fly, similarly to~\cite{Andreas2016}, where they
tried to produce unambiguous captions for clip-art images. The generation model generates some candidate expressions and passes them through the comprehension model. The final output expression is the one with highest generation-comprehension score which we will describe later.

The {\bf training by proxy} method is closer in spirit to GANs. The generation and
comprehension model are connected and the generation model is
optimized to lower discriminative comprehension loss (in addition to
the cross-entropy loss). We investigate several training strategies for this method and a trick to make proxy model trainable by standard back-propagation.  Compared to generate-and-rerank method, the training by proxy method doesn't require additional region proposals during test time.



%% file: related.tex
\section{Related work}\label{sec:related}

The main approach in modern image captioning
literature~\cite{Vinyals2014,Karpathy2015,Ma2016} is to encode an
image using a convolutional neural network (CNN), and then feed this as input to an RNN, which is able to generate a arbitrary-length sequence of words.

While captioning typically aims to describe an entire image, some work
takes regions into consideration, by incorporating them in an
attention mechanism~\cite{Xu2015,Liu2016}, alignment of words/phrases within
sentences to regions~\cite{Karpathy2015}, or by defining ``dense'' captioning on
a per-region basis~\cite{Johnson2015}. The latter includes a dataset of captions collected without requirement to be unambiguous, so they cannot be regarded as referring expression.

Text-based image retrieval has been considered as a task relying on
image
captioning~\cite{Vinyals2014,Karpathy2015,Ma2016,Xu2015}. However, it
can also be regarded as a multi-modal embedding
task. In previous works~\cite{frome2013devise,Wang2016a,WSABIE} such
embeddings have been trained separately for visual and textual input,
with the objective to minimize matching loss, e.g., hinge loss on cosine
distance, or to enforce partial order on 
captions and images~\cite{Vendrov2015}. \cite{Reed2016} tried
to retrieve from fine-grained images given a text, where they explored
different network structures for text embedding. 


Closer to the focus of this paper, referring expressions have
attracted interest after the release of the standard
datasets~\cite{Kazemzadeh2014,Yu2016,Mao2016a}. In~\cite{Hu2015} a caption
generation model is appropriated for a generation task, by evaluating
the probability of a sentence given an image $P(S|I)$ as the matching
score. Concurrently,~\cite{Mao2016} at the same time proposed a joint
model, in which comprehension and generation aspects are trained using
max-margin Maximum Mutual Information (MMI) training. Both papers used
whole image, region and location/size features. Based on the model
in~\cite{Mao2016a}, both~\cite{Nagaraja2016} and~\cite{Yu2016a}
try to model context regions in their frameworks. 

Our method is trying to combine simple models and replace the max margin
loss, which is orthogonal to modeling context, with a surrogate closer
to the eventual goal -- human comprehension. This requires a
comprehension model, which, given a referring expression, infers the appropriate region in the image.

Among comprehension models proposed in literature,~\cite{Rohrbach2015}
uses multi-modal embedding and sets up the comprehension task as a multi-class classification. Later,~\cite{Fukui2016} achieves a slight improvement by replacing the concatenation layer with a compact bilinear pooling layer. The comprehension model used in this paper belongs to this multi-modal embedding category.

\begin{figure}[t]
\centering

\begin{minipage}[b]{0.48\textwidth}
\centering
\includegraphics[width=\textwidth]{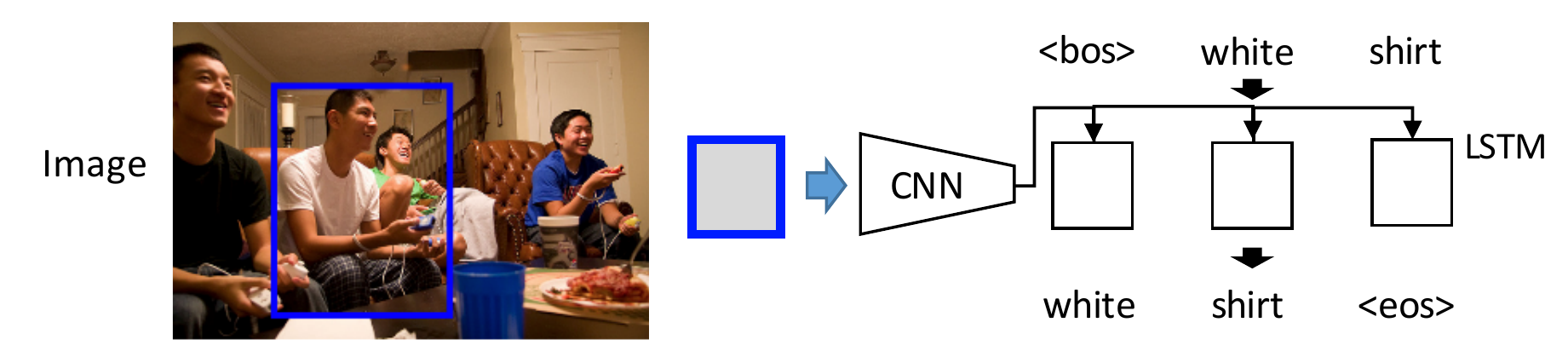}
\caption{Illustration of how the generation model describes region inside the blue bounding box. $<$bos$>$ and $<$eos$>$ stand for beginning and end of sentence.}
\label{fig:generation} 
\end{minipage}
\begin{minipage}[b]{0.48\textwidth}
\centering
\includegraphics[width=\textwidth]{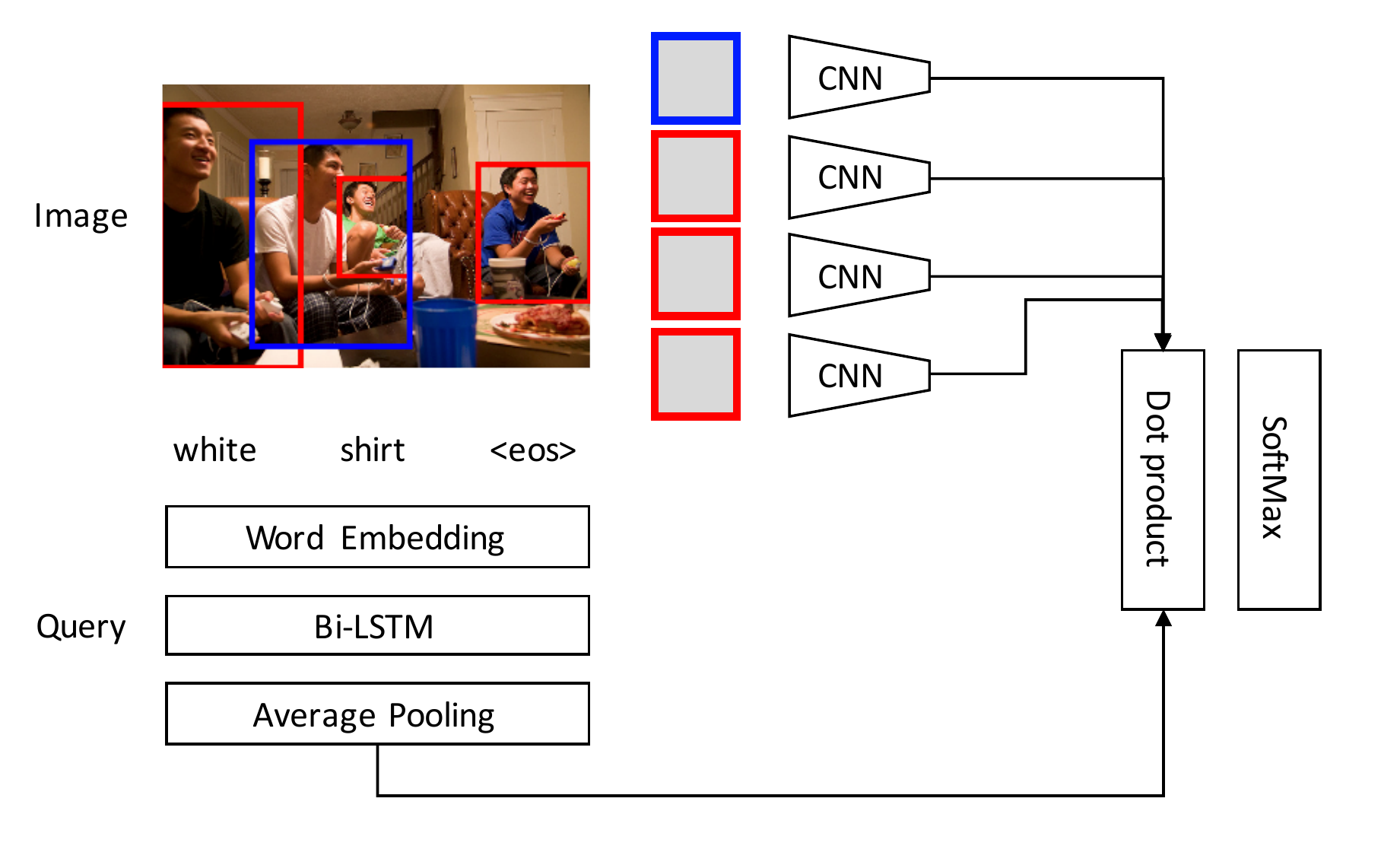}
\caption{Illustration of comprehension model using softmax loss. The blue bounding box is the target region, and the red ones are incorrect regions. The CNNs share the weights.}
\label{fig:comprehension} 
\par\vspace{0pt}
\end{minipage}
\end{figure}

The ``speaker-listener'' model in~\cite{Andreas2016} attempts to produce discriminative captions that
can tell images apart. The speaker is trained to generate captions, and a
listener to prefer the correct image over a wrong one, given the
caption. At test time, the listener reranks the captions sampled
from the speaker. Our generate-and-rerank method is based on translating this idea to referring expression generation.

%% file: models.tex
\section{Generation and comprehension models}\label{sec:model}
We start by defining the two modules used in the collaborative
architecture we propose. Each of these can be trained as a standalone
machine for the task it solves, given a data set with ground truth
regions/referring expressions. 

\subsection{Expression generation model}\label{sec:generation}
We use a simple expression generation model introduced
in~\cite{Yu2016,Mao2016a}. The generation task takes inputs of an image $I$ and an internal region $r$, and outputs an expression $w$.
\begin{align}
G:I\times r \rightarrow w
\end{align}
To address this task, we build a model $P_G(w|I,r)$. With this $P_G$, we have 
\begin{align}
G(I, r) = \argmax_w P_G(w | I,r)
\end{align}

To train $P_G$, we need a set of image, region and expression tuples, $\{(I_i, w_i, r_i)\}$. We can then train this model by maximizing the likelihood
\begin{align}
P_G^* = \argmax_{P_G} \sum_i \log P_G(w_i|I_i, r_i)
\end{align}

Specifically, the generation model is an encoder-decoder network.
First we need to encode the visual information from $r_i$ and $I_i$. As
in~\cite{Hu2015,Yu2016,Nagaraja2016}, we use such representation: target
object representation $o_i$, global context feature $g_i$ and
location/size feature $l_i$. In our experiments, $o_i$ is the
activation on the cropped region $r_i$ of the last fully connected layer
{\tt fc7} of VGG-16~\cite{simonyan2014very}; $g_i$ is the {\tt fc7} activation on the
whole image $I_i$; $l_i$ is a 5D vector encoding the opposite
corners of the bounding box of $r_i$, as well as the bounding box
size relative to the image size. The final visual feature $v_i$ of the
region is an affine transformation of the concatenation of three features:
\begin{equation}
v_i = W[o_i, g_i, l_i] + b
\end{equation}

To generate a sequence we
use a uni-directional LSTM decoder\cite{hochreiter1997long}. Inputs of LSTM at each time step include
the visual features and the previous word embedding. The output of the
LSTM at a time step is the distribution of predicted next word. Then the whole model is trained to minimize cross entropy loss, which is equivalent to maximize the likelihood.
\begin{align}
  \label{eq:celoss}
L_{gen} &=  \sum_{i}\sum_{t = 1}^T \log P_G(w_{i,t}|w_{i,<t}, I_i, r_i),\\
P_G^* &= \argmin_{P_G} L_{gen},
\end{align}
where , $w_{i, t}$ is the t-th word of  
ground truth expression $w_i$, and $T$ is the length of $w_i$.

In practice, instead of precisely inferring the $\argmax_w P_G(w | I,r)$, people use beam search, greedy search or sampling to get the output.

Figure~\ref{fig:generation} shows the structure of our generation model.

\subsection{Comprehension}
The comprehension task is to select a region (bounding box) $\hat r$ from a
set of regions  $\calR = \{r_i\}$ given a query expression $q$ and the image $I$.
\begin{align}
C: I\times q\times \calR \rightarrow r,\  r\in \calR
\end{align}
We also define the comprehension model as a posterior distribution $P_C(r|I, q, \calR)$.
The estimated region given a comprehension model is: $\hat r = \argmax_r P_C(r|I, q, \calR)$.

In general, our comprehension model is very similar to~\cite{Rohrbach2015}. To build the model, we first define a similarity function $f_{sim}$.
We use the same visual feature encoder structure as in generation model. For the query expression, we use a one-layer bi-directional LSTM~\cite{Graves2013} to encode it. We take the averaging over the hidden vectors of each timestep so that we can get a fixed-length representation for an arbitrary length of query. 
\begin{align}
h = f_{LSTM}(\E \Q),
\end{align}
where $\E$ is the word embedding matrix initialized from pretrained word2vec\cite{mikolov2013efficient} and $\Q$ is a one-hot representation of the query expression, i.e. $\Q_{i,j} = \1(q_i = j)$. 

Unlike \cite{Rohrbach2015}, which uses concatenation + MLP to
calculate the similarity, we use a simple dot product as in \cite{Dhingra2016}.
\begin{align}
f_{sim}(I, r_i, q) = v_i^T * h.
\end{align}

We consider two formulations of the comprehension
task as classification. The
per-region logistic loss 
\begin{align}
  \label{eq:lbin}
  &P_C(r_i|I, q)=\sigma(f_{sim}(I, r_i, q)),\\
  &L_{bin}=-\log P_C(r_{j^\ast}|I, q) - \sum_{i \neq j^\ast}\log (1- P_C(r_i|I, q)),
\end{align}
where $r_{j^\ast}$ is ground truth region, corresponds to a per-region
classification: is this region the right match for the expression or not.  The softmax loss
\begin{align}
  \label{eq:lmulti}
  P_C(r_i|I, q, \calR)&=\frac{e^{s_i}}{\sum_i e^{s_i}},\\
  L_{multi}&=-\log P_C(r_{j^\ast}|I, q, \calR),
\end{align}
where $s_i =
f_{sim}(I, r_i, q)$, frames the task as a multi-class
classification: which region in the set should be matched to the
expression. 

The model is trained to minimize the comprehension loss.
$P_C^* = \argmin_{P_C} L_{com}$, where $L_{com}$ is either $L_{bin}$ or $L_{multi}$.

Figure~\ref{fig:comprehension} shows the structure of our generation model under multi-class classification formulation.

%% file: methods.tex
\section{Comprehension-guided generation}\label{sec:method}
Once we have trained the comprehension model, we can start using it as
a proxy for human comprehension, to guide expression generator. Below
we describe two such approaches: one is applied at training time, and
the other at test time. 

\newcommand{\propo}{\mathcal{R}}

\subsection{Training by proxy}\label{sec:proxy}
Consider a referring expression generated by $G$ for a given training example
of an image/region pair $(I,r)$. The generation loss $L_{gen}$ will inform the
generator how to modify its model to maximize the probability of the
ground truth expression $w$. The comprehension model $C$ can provide an
alternative, complementary signal: how to modify $G$ to maximize
the discriminativity of the generated expression, so that $C$ selects
the correct region $r$ among the proposal set
$\propo$. Intuitively, this signal should push down on probability of
a word if it's unhelpful for comprehension, and pull that probability
up if it is helpful. 



Ideally, we hope to minimize the comprehension loss
of the output of the generation model $L_{com}(r|I, \calR, \tilde \Q)$,
where $\tilde \Q$ is the 1-hot encoding of $\tilde q = G(I, r)$, 
with $K$ rows (vocabulary size) and $T$ columns (sequence length).

We hope to update the generation model according to the gradient of loss with respect to the model parameter $\theta_G$. By chain rule, 
\begin{align}
\frac{\partial L_{com}}{\partial \theta_G} = 
\frac{\partial L_{com}}{\partial \tilde \Q}\frac{\partial \tilde \Q}{\partial \theta_G}
\end{align}
However, $\tilde \Q$ is inferred by some algorithm which is not differentiable. To address this issue, \cite{Ranzato2016,Bahdanau2016,Yu2016a} applied reinforcement
learning methods. However, here we use an approximate method borrowing from the idea of soft attention mechanism~\cite{Xu2015, DzmitryBahdana2014}.

We define a matrix $\BP$ which has the same size as $\tilde \Q$. The i-th column of $\BP$ is -- instead of the one-hot vector of the generated word $i$ -- the distribution of the i-th word produced by $P_G$, i.e. 
\begin{align}
\BP_{i,j} = P_G(w_i = j).
\end{align}
$\BP$ has several good properties. First, $\BP$ has the same size as
$\tilde \Q$, so that the we can still compute the query feature by
replacing the $\tilde \Q$ by $\BP$, i.e. $h =
f_{LSTM}(\E\BP)$. Secondly, the sum of each column in $\BP$ is 1, just
like $\tilde \Q$. Thirdly, $\BP$ is differentiable with respect to
generator's parameters.

Now, the gradient of $\theta_G$ is calculated by:
\begin{align}\label{eq:diffapprox}
\frac{\partial L_{com}}{\partial \theta_G} = 
\frac{\partial L_{com}}{\partial \BP}\frac{\partial \BP}{\partial \theta_G}
\end{align}

We will use this approximate gradient in the following three methods.
\subsubsection{Compound loss}\label{sec:CL}


Here we introduce how we integrate the comprehension model to guide the training of the generation model.

The cross-entropy loss~\eqref{eq:celoss}
encourages fluency of the generated expression, but disregards its
discriminativity. We address this by using the comprehension model as
a source of an additional loss signal. Technically, we define a
compound loss 
\begin{equation}\label{eq:lcompound}
L = L_{gen} + \lambda L_{com} 
\end{equation}
where the comprehension loss $L_{com}$ is either the
logistic~\eqref{eq:lbin} or the softmax~\eqref{eq:lmulti} loss; the
balance term $\lambda$ determines the relative importance of fluency
vs. discriminativity in $L$. 

Both $L_{gen}$ and $L_{com}$ take as input $G$'s distribution over the
$i$-th word $P_G(w_i|I,r, w_{<i})$, where
the preceding words $w_{<i}$ are from the ground truth expression.

Replacing $\tilde{\mathbf{Q}}$ with $\mathbf{P}$
(Sec.~\ref{sec:proxy}) allows us to
train the model by back-propogation from the compound loss~\eqref{eq:lcompound}.

\subsubsection{Modified Scheduled sampling training}
Our final goal is to generate comprehensible expression during test time. However, in compound loss, the loss is calculated given the ground truth input while during test time each token is generated by the model, thus yielding a discrepancy between how the model is used during training and at test time. Inspired by similar motivation, \cite{Bengio2015} proposed scheduled sampling which allows the model to be trained with a mixture of ground truth data and predicted data. Here, we propose this modified schedule sampling training to train our model.

During training, at each iteration $i$, before forwarding through the
LSTMs, we draw a random variable $\alpha$ from a Bernoulli
distribution with probability $\epsilon_i$. If $\alpha = 1$, we feed
the ground truth expression to LSTM frames, and minimize cross entropy loss. If
$\alpha = 0$, we sample the whole sequence step by step according to
the posterior, and the input of comprehension model is $P_G(w_i|I, r,
\hat w_{<i})$, where $\hat w_{<i}$ are the \emph{sampled} words. We update
the model by minimizing the comprehension loss. Therefore, $\alpha$
serves as a dispatch mechanism, randomly alternating between the
sources of data for the LSTMs and the components of the compound loss.


We start the modified scheduled sampling training from a pretrained generation model trained on cross entropy loss using the ground truth sequences. As the training progresses, we linearly decay $\epsilon_i$ until a preset minimum value $\epsilon$. The minimum probability prevents the model from degeneration. If we don't set the minimum, when $\epsilon_i$ goes to 0, the model will lose all the ground truth information, and will be purely guided by the comprehension model. 
This would lead the generation model to discover those pathological optimas that exist in neural classification models\cite{goodfellow2014explaining}. In this case, the generated expressions would do ``well" on comprehension model, but no longer be intelligible to human. See Algorithm \ref{alg:mss} for the pseudo-code.

\makeatletter
\def\BState{\State\hskip-\ALG@thistlm}
\makeatother

\begin{algorithm}
\caption{Modified scheduled sampling training}\label{alg:mss}
\begin{algorithmic}[1]
\State Train the generation model $G$.
\State Set the offset $k$ ($0\leq k\leq 1$), the slope of decay $c$, minimum probability $\epsilon$, number of iterations $N$.
\For{$i=1, N$}
\State $\epsilon_i \gets \max(\epsilon, k - ci)$
\State Get a sample from training data, $(I, r, w)$
\State Sample the $\alpha$ from Bernoulli distribution, where $P(\alpha = 1) = \epsilon_i$
\If {$\alpha = 1$}
\State Minimize $L_{gen}$ with the ground truth input.
\Else
\State Sample a sequence $\hat w$ from $P_G(w|I, r)$
\State Minimize $L_{com}$ with the input $P_G(w_j|I, r, \hat w_{<j})$, $j \in [1, T]$
\EndIf
\EndFor

\end{algorithmic}
\end{algorithm}



\subsubsection{Stochastic mixed sampling}

Since modified scheduled sampling training samples a whole sentence at a time, it would be hard to get useful signal if there is an error at the beginning of the inference. We hope to find a method that can slowly deviate from the original model and explore.

Here we borrow the idea from mixed incremental cross-entropy reinforce(MIXER)\cite{Ranzato2016}. Again, we start the model from a pretrained generator. Then we introduce model predictions during training with an annealing schedule so as to gradually teach the model to produce stable sequences. For each iteration $i$, We feed the input for the first $s_i$ steps, and sample the rest $T - s_i$ words, where $0\leq s_i \leq T$, and $T$ is the maximum length of expressions. We define $s_i = s + \Delta s$, where $s$ is a base step size which gradually decreases during training, and $\Delta s$ is a random variable which follows geometric distribution: $P(\Delta s = k) = (1-p)^{k+1}p$. This $\Delta s$ is the difference between our method and MIXER. We call this method: Stochastic mixed incremental cross-entropy comprehension(SMIXEC).

By introducing this term $\Delta s$, we can control how much supervision we want to get from ground truth by tuning the value $p$. This is also for preventing the model from producing pathological optimas. Note that, when $p$ is 0, $\Delta s$ will always be large enough so that it's just cross entropy loss training. When $p$ is 1, $\Delta s$ will always equal to 0, which is equivalent to MIXER annealing schedule. See Algorithm \ref{alg:smixec} for the pseudo-code.

\begin{algorithm}
\caption{Stochastic mixed incremental cross-entropy comprehension (SMIXEC)}\label{alg:smixec}
\begin{algorithmic}[1]
\State Train the generation model $G$.
\State Set the geometric distribution parameter $p$, maximum sequence length $T$, period of decay $d$, number of iterations $N$.
\For{$i=1, N$}
\State $s \gets \max(0, T - \lceil i / d\rceil)$
\State Sample $\Delta s$ from geometric distribution with success probability $p$
\State $s_i \gets \min(T, s + \Delta s)$
\State Get a sample from training data, $(I, r, w)$
\State Run the $G$ with ground truth input in the first $s_i$ steps, and sampled input in the remaining $T - s_i$
\State Get $L_{gen}$ on first $s_i$ steps, and $L_{com}$ on whole sentence but with input $\{w_{1\ldots s_i}, P_G(w_{s_i+1 \ldots T}|I, r, w_{1\cdots s_i}, \hat w_{s_i+1 \ldots T})\}$

\State Minimize $L_{com} + \lambda L_{gen}$. (Not backprop through $w_{1\ldots s_i}$)
\EndFor

\end{algorithmic}
\end{algorithm}



\subsection{Generate-and-rerank}

Here we propose a different strategy to generate better expressions. Instead of using comprehension model for training a generation model, we compose the comprehension model during test time. The pipeline is similar to \cite{Andreas2016}.

Unlike in Sec.~\ref{sec:generation}, we not only need image $I$ and region $r$ as input, but also a region set $\calR$. Suppose we have a generation model and a comprehension model which are trained pretrained. The steps are as follows:
\begin{enumerate}
\itemsep=-0pt
\item Generate candidate expressions $\{c_1, \ldots, c_n \}$ according to $P_G(\cdot|I,r)$.
\item Select $c_k$ with $k = \argmax_i score(c_i)$.
\end{enumerate}
Here, we don't use beam search because we want the candidate set to be more diverse. And we define the score function as a weighted combination of the log perplexity and comprehension loss (we assume to use softmax loss here).
\begin{align}
	score(c) = \frac{1}{T} \sum_{k = 1}^{T} \log p_G(c_k|r, c_{1..k-1}) \nonumber\\
	 + \gamma \log p_C(r|I, \calR, c),
\end{align}
where $c_k$ is the k-th token of $c$, $T$ is the length of $c$.

This can be viewed as a weighted joint log probability that an expression to be both nature and unambiguous. The log perplexity term ensures the fluency, and the comprehension loss ensures the chosen expression to be discriminative.

%% file: experiments.tex
\begin{table*}[ht]\small
  \centering
    \begin{tabular}{|l|c|c|c|c|c|c|c|c|c|c|}
    \hline
    \multirow{3}[6]{*}{} & \multicolumn{4}{c|}{RefCOCO}  & \multicolumn{4}{c|}{RefCOCO+} & \multicolumn{2}{c|}{RefCOCOg} \bigstrut\\
\cline{2-11}          & \multicolumn{2}{c|}{Test A} & \multicolumn{2}{c|}{Test B} & \multicolumn{2}{c|}{Test A} & \multicolumn{2}{c|}{Test B} & \multicolumn{2}{c|}{Val} \bigstrut\\
\cline{2-11}          & GT    & DET   & GT    & DET   & GT    & DET   & GT    & DET   & GT    & DET \bigstrut\\
    \hline
    MLE\cite{Yu2016} & 63.15\% & 58.32\% & 64.21\% & 48.48\% & 48.73\% & 46.86\% & 42.13\% & 34.04\% & 55.16\% & 40.75\% \bigstrut\\
    \hline
    MMI\cite{Yu2016}   & 71.72\% & 64.90\% & 71.09\% & 54.51\% & 52.44\% & 54.03\% & 47.51\% & 42.81\% & 62.14\% & 45.85\% \bigstrut\\
    \hline
    visdif+MMI\cite{Yu2016} & 73.98\% & 67.64\% & 76.59\% & 55.16\% & 59.17\% & 55.81\% & \textbf{55.62\%} & 43.43\% & 64.02\% & 46.86\% \bigstrut\\
    \hline
    Neg Bag\cite{Nagaraja2016} & \textbf{75.6\%} & 58.6\% & \textbf{78.0\%} & \textbf{56.4\%} & -     & -     & -     & -     & \textbf{68.4\%} & 39.5\% \bigstrut\\
    \hline
    \hline
    Ours  & 74.14\% & \textbf{68.11\%} & 71.46\% & 54.65\% & 59.87\% & 56.61\% & 54.35\% & \textbf{43.74\%} & 63.39\% & 47.60\% \bigstrut\\
    \hline
    Ours(w2v) & 74.04\% & 67.94\% & 73.43\% & 55.18\% & \textbf{60.26\%} & \textbf{57.05\%} & 55.03\% & 43.33\% & 65.36\% & \textbf{49.07\%} \bigstrut\\
    \hline
    \end{tabular}%
  \caption{Comprehensions results on RefCOCO, RefCOCO+, RefCOCOg
    datasets. GT: the region set contains ground truth bounding boxes; DET: region set contains proposals generated from detectors. w2v means initializing the embedding layer using pretrained word2vec.}      
   \label{tab:comprehension}

\end{table*}%

\begin{table}[thbp]\small
  \centering
    \begin{tabular}{|l|c|}
    \hline
          & RefCLEF Test \bigstrut\\
    \hline
    SCRC\cite{Hu2015}  & 17.93\% \bigstrut\\
    \hline
    GroundR\cite{Rohrbach2015} & 26.93\% \bigstrut\\
    \hline
    MCB\cite{Fukui2016}   & 28.91\% \bigstrut\\
    \hline
    \hline
    Ours  & 31.25\% \bigstrut\\
    \hline
    Ours(w2v) & \textbf{31.85\%} \bigstrut\\
    \hline
    \end{tabular}%
  \caption{Comprehension on RefClef (EdgeBox proposals)}  \label{tab:compre_clef}
\end{table}%

\section{Experiments}\label{sec:experiment}

We base our experiments on the following data sets.

{\bf RefClef(ReferIt)}\cite{Kazemzadeh2014} contains 20,000 images from IAPR TC-12 dataset\cite{Grubinger2006}, together with segmented image regions from SAIAPR-12 dataset\cite{Escalante2010}. The dataset is split into 10,000 for training and validation and 10,000 for test. There are 59,976 (image, bounding box, description) tuples in the trainval set and 60,105 in the test set.

{\bf RefCOCO(UNC RefExp)}\cite{Yu2016} consists of 142,209 referring expressions for
50,000 objects in 19,994 images from COCO\cite{lin2014microsoft}, collected using the ReferitGame~\cite{Kazemzadeh2014}

{\bf RefCOCO+}\cite{Yu2016} has 141,564 expressions for 49,856 objects in
19,992 images from COCO. RefCOCO+ dataset players are disallowed from using
location words, so this dataset focuses more on purely appearance based description.

{\bf RefCOCOg(Google RefExp)}\cite{Mao2016a} consists of 85,474
referring expressions for 54,822 objects in 26,711 images from COCO; it contains
longer and more flowery expressions than RefCOCO and RefCOCO+.

\subsection{Comprehension}

We first evaluate our comprehension model on human-made expressions, to assess its ability to provide useful signal. 

There are two comprehension experiment settings as in
\cite{Mao2016a,Yu2016,Nagaraja2016}. First, the input region set
$\calR$ contains only ground truth bounding boxes for objects, and a
hit is defined by the model choosing the correct region the expression
refers to. In the second setting, $\calR$ contains proposal regions
generated by FastRCNN detector\cite{girshick2015fast}, or by other
proposal generation
methods\cite{edge-boxes-locating-object-proposals-from-edges}. Here a
hit occurs when the model chooses a proposal with intersection over
union(IoU) with the ground truth of 0.5 or higher. We used precomputed proposals from \cite{Yu2016,Mao2016a,Hu2015} for all four datasets.

In RefCOCO and RefCOCO+, we have two test sets: testA contains people and testB contains all other objects. For RefCOCOg, we evaluate on the validation set. For RefClef, we evaluate on the test set.

We train the model using Adam optimizer~\cite{kingma2014adam}. The word embedding size is 300, and the hidden size of bi-LSTM is 512. The length of visual feature is 1024. For RefCOCO, RefCOCO+ and RefCOCOg, we train the model using softmax loss, with ground truth regions as training data. For RefClef dataset, we use the logistic loss. The training regions are composed of ground truth regions and all the proposals from Edge Box~\cite{edge-boxes-locating-object-proposals-from-edges}. The binary classification is to tell if the proposal is a hit or not.

\begin{table*}[thbp!]\small
  \centering
    \begin{tabular}{|l|c|c|c|c|c|c|c|c|c|c|}
    \multicolumn{11}{c}{RefCOCO} \bigstrut[b]\\
    \hline
    \multirow{2}[4]{*}{} & \multicolumn{5}{c|}{Test A}           & \multicolumn{5}{c|}{Test B} \bigstrut\\
\cline{2-11}          & Acc   & BLEU 1 & BLEU 2 & ROUGE & METEOR & Acc   & BLEU 1 & BLEU 2 & ROUGE & METEOR \bigstrut\\
    \hline
    MLE\cite{Yu2016} & 74.80\% & 0.477 & 0.290 & 0.413 & 0.173 & 72.81\% & 0.553 & \textbf{0.343} & 0.499 & 0.228 \bigstrut\\
    \hline
    MMI\cite{Yu2016}   & 78.78\% & 0.478 & \textbf{0.295} & 0.418 & 0.175 & 74.01\% & 0.547 & 0.341 & 0.497 & 0.228 \bigstrut\\
    \hline
    CL    & \textbf{80.14\%} & 0.4586 & 0.2552 & 0.4096 & 0.178 & 75.44\% & 0.5434 & 0.3266 & \textbf{0.5056} & \textbf{0.2326} \bigstrut\\
    \hline
    MSS   & 79.94\% & 0.4574 & 0.2532 & 0.4126 & 0.1759 & \textbf{75.93\%} & 0.5403 & 0.3232 & 0.5010 & 0.2297 \bigstrut\\
    \hline
    SMIXEC & 79.99\% & \textbf{0.4855} & 0.2800 & \textbf{0.4212} & \textbf{0.1848} & 75.60\% & \textbf{0.5536} & 0.3426 & 0.5012 & 0.2320 \bigstrut\\
    \hline
    \hline
    MLE+sample & 78.38\% & 0.5201 & \textbf{0.3391} & 0.4484 & 0.1974 & 73.08\% & 0.5842 & 0.3686 & 0.5161 & 0.2425 \bigstrut\\
    \hline
    Rerank & \textbf{97.23\%} & \textbf{0.5209} & \textbf{0.3391} & \textbf{0.4582} & \textbf{0.2049} & \textbf{94.96\%} & \textbf{0.5935} & \textbf{0.3763} & \textbf{0.5259} & \textbf{0.2505} \bigstrut\\
    \hline
    \multicolumn{11}{c}{RefCOCO+} \bigstrut\\
    \hline
    \multirow{2}[4]{*}{} & \multicolumn{5}{c|}{Test A}           & \multicolumn{5}{c|}{Test B} \bigstrut\\
\cline{2-11}          & Acc   & BLEU 1 & BLEU 2 & ROUGE & METEOR & Acc   & BLEU 1 & BLEU 2 & ROUGE & METEOR \bigstrut\\
    \hline
    MLE\cite{Yu2016} & 62.10\% & \textbf{0.391} & \textbf{0.218} & \textbf{0.356} & 0.140 & 46.21\% & 0.331 & 0.174 & 0.322 & 0.135 \bigstrut\\
    \hline
    MMI\cite{Yu2016}   & 67.79\% & 0.370 & 0.203 & 0.346 & 0.136 & 55.21\% & 0.324 & 0.167 & 0.320 & 0.133 \bigstrut\\
    \hline
    CL    & 68.54\% & 0.3683 & 0.2041 & 0.3386 & 0.1375 & \textbf{55.87\%} & \textbf{0.3409} & \textbf{0.1829} & \textbf{0.3432} & \textbf{0.1455} \bigstrut\\
    \hline
    MSS   & \textbf{69.41\%} & 0.3763 & 0.2126 & 0.3425 & 0.1401 & 55.59\% & 0.3386 & 0.1823 & 0.3365 & 0.1424 \bigstrut\\
    \hline
    SMIXEC & 69.05\% & 0.3847 & 0.2125 & 0.3507 & \textbf{0.1436} & 54.71\% & 0.3275 & 0.1716 & 0.3194 & 0.1354 \bigstrut\\
    \hline
    \hline
    MLE+sample & 62.45\% & 0.3925 & 0.2256 & 0.3581 & 0.1456 & 47.86\% & 0.3354 & 0.1819 & 0.3370 & 0.1470 \bigstrut\\
    \hline
    Rerank & \textbf{77.32\%} & \textbf{0.3956} & \textbf{0.2284} & \textbf{0.3636} & \textbf{0.1484} & \textbf{67.65\%} & \textbf{0.3368} & \textbf{0.1843} & \textbf{0.3441} & \textbf{0.1509} \bigstrut\\
    \hline
    \end{tabular}
  \caption{Expression generation evaluated by automated metrics. Acc:
    accuracy of the trained comprehension model on generated
    expressions. We separately mark in bold the best results for
    single-output methods (top) and sample-based methods (bottom) that
  generate multiple expressions and select one.}
  \label{tab:generation}
\end{table*}%

\begin{figure*}[hbt]
\centering
\includegraphics[width=\textwidth]{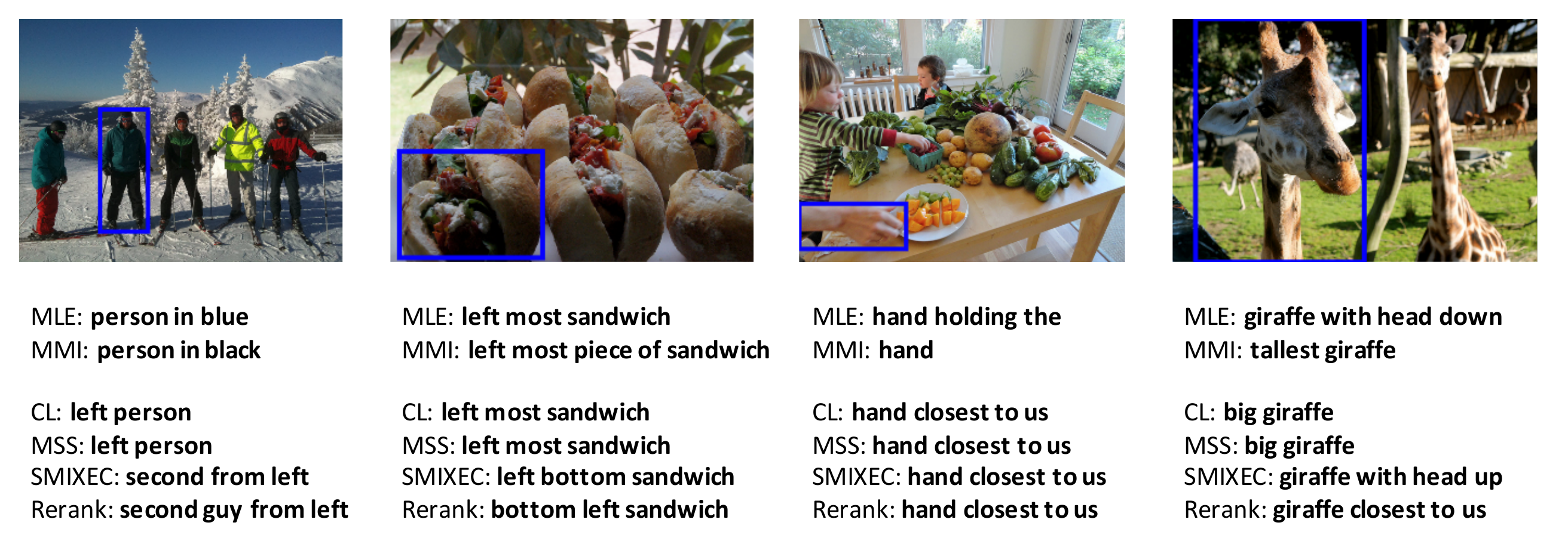}
\caption{Generation results. The images are from RefCOCO testA, RefCOCO testB, RefCOCO+ testA and RefCOCO+ testB (from left to right).}
\label{fig:gen_result}
\end{figure*}

Table \ref{tab:comprehension} shows our results on RefCOCO, RefCOCO+
and RefCOCOg compared to recent algorithms. Among these, MMI
represents Maximum Mutual Information which uses max-margin loss to
help the generation model better comprehend. With the same visual
feature encoder, our model can get a better result compared to MMI in
\cite{Yu2016}. Our model is also competitive with recent, more complex
state-of-the-art models~\cite{Yu2016,Nagaraja2016}. Table
\ref{tab:compre_clef} shows our results on RefClef where we only test
in the second setting to compare to existing results; our model, which
is a modest modification of~\cite{Rohrbach2015}, obtains state of the art
accuracy in this experiment.
\subsection{Generation}

We evaluate our expression generation methods, along with baselines, on RefCOCO and RefCOCO+. Table~\ref{tab:generation} shows our evaluation of different methods based on automatic caption generation metrics. We also add an `Acc' column, which is the ``comprehension accuracy" of the generated expressions according to our comprehension model: how well our comprehension model can comprehend the generated expressions.

The two baseline models are max likelihood(MLE) and maximum mutual information(MMI) from \cite{Yu2016}. Our methods include compound loss(CL), modified scheduled sampling(MSS), stochastic mixed incremental cross-entropy comprehension(SMIXEC) and also generate-and-rerank(Rerank). And the MLE+sample is designed for better analyzing rerank model. 

For the two baseline models and our three strategies for training by proxy
 method, we use greedy search to generate an expression. The MLE+sample
and Rerank methods generate an expression by choosing a best one from 100 sampled expressions.

Our generate-and-rerank (Rerank in Table~\ref{tab:generation}) model gets consistently better results on automatic comprehension accuracy and on fluency-based metrics like
BLEU. To see if the improvement is from sampling or reranking, we also
sampled 100 expressions on MLE model and choose the one with
the lowest perplexity (MLE+sample in Table~\ref{tab:generation}). The
generate-and-rerank method still has better results, showing benefit
from comprehension-guided reranking.

We can see that our training by proxy method can get higher
accuracy under the comprehension model. This confirms the
effectiveness of our collaborative training by proxy method, despite
the differentiable approximation~\eqref{eq:diffapprox}.

Among the three training schedules of training by proxy, there is no clear winner. In RefCOCO, our SMIXEC method outperforms basic MMI method with higher comprehension accuracy and higher caption generation metrics. The compound loss and modified scheduled sampling seem to suffer from optimizing over the accuracy. However, in RefCOCO+, our three models seem to perform very differently. The compound loss works better on TestB; the SMIXEC works best on TestA and the MSS method works reasonably well on both. We currently don't have a concrete explanation why this is happening.
\paragraph{Human evaluations} From \cite{Yu2016}, we know the human
evaluations on the expressions are not perfectly correlated with
language-based caption metrics. Thus, we performed human evaluations
on the expression generation for 100 images randomly chosen from each
split of RefCOCO and RefCOCO+. Subjects clicked onto the object which
they thought was the most probable match for a generated expression. Each
image/expression example was presented to two subjects, with a hit
recorded only when both subjects clicked inside the correct region.

\begin{table}[tbph]
  \centering
  \caption{Human evaluations}
    \begin{tabular}{|l|c|c|c|c|}
    \hline
    \multirow{2}[4]{*}{} & \multicolumn{2}{c|}{RefCOCO} & \multicolumn{2}{c|}{RefCOCO+} \bigstrut\\
\cline{2-5}          & Test A & Test B & TestA & TestB \bigstrut\\
    \hline
    MMI\cite{Yu2016}   & 53\%  & 61\%  & 39\%  & 35\% \bigstrut\\
    \hline
    SMIXEC & 62\%  & 68\%  & \textbf{46\%} & 25\% \bigstrut\\
    \hline
    Rerank & \textbf{66\%} & \textbf{75\%} & 43\%  & \textbf{47\%} \bigstrut\\
    \hline
    \end{tabular}%
  \label{tab:generation_human}%
\end{table}%

The results from human evaluations with MMI, SMIXEC and our
generate-and-rerank method are in
Table~\ref{tab:generation_human}. On RefCOCO, both of our
comprehension-guided methods appear to generate better (more
informative) referring expressions. 
On RefCOCO+, the result are similar to those on RefCOCO on TestA, but our training by proxy methods
performs less well on TestB. 

Fig \ref{fig:gen_result} shows some example generation results on test images.

%% file: conclusion.tex
\section{Conclusion}\label{sec:end}
In this paper, we propose to use learned comprehension models to guide
generating better referring expressions. Comprehension guidance can be
incorporated at training time, with a training by proxy method, where
the discriminative comprehension loss (region retrieval based on
generated referring expressions) is included in training the
expression generator. Alternatively comprehension guidance can be used
at
test time, with a generate-and-rerank method which uses model
comprehension score to select among
multiple proposed expressions. Empirical evaluation
shows both to be promising, with the generate-and-rerank method
obtaining particularly good results across data sets.

Among directions for future work we are interested to explore 
alternative training regimes, in particular an adaptation of the GAN
protocol to referring expression generation. We will try to
incorporate context objects (other regions in the image) into
representation for a reference region. Finally, while at the moment
the generation and comprehension models are completely separate, it is
interesting to consider weight sharing. 

%% file: acknowledge.tex
\section*{Acknowledgements}

We thank Licheng Yu for providing the baseline code. We also thank those who participated in evaluations. Finally, we gratefully acknowledge the support of NVIDIA Corporation with the donation of GPUs used for this research.